\documentclass[conference]{IEEEtran}
\usepackage{graphicx}
\usepackage{epstopdf}
\ifCLASSINFOpdf
\else
\fi
\usepackage{amsmath}
\usepackage{amssymb}
\usepackage{hyperref}
\hypersetup{linkcolor=black,anchorcolor=black,citecolor=black,filecolor=black,menucolor=black,urlcolor=black,colorlinks=true}

\begin{document}
%
\title{Signal reconstruction via operator guiding}

\author{\IEEEauthorblockN{Andrew Knyazev}
\IEEEauthorblockA{Mitsubishi Electric Research Laboratories (MERL)\\
  201 Broadway, 8th floor\\
  Cambridge, MA 02139-1955 \\
  Email: knyazev@merl.com}
\and
\IEEEauthorblockN{Alexander Malyshev}
\IEEEauthorblockA{University of Bergen \\
  Department of Mathematics \\
  Postbox 7803, 5020 Bergen, Norway \\
  Email: alexander.malyshev@uib.no}
}


%


\maketitle

\begin{abstract}
Signal reconstruction from a sample using an orthogonal projector onto a guiding subspace is theoretically well justified, but may be difficult to practically implement. We~propose more general guiding operators, which increase signal components in the guiding subspace relative to those in a complementary subspace, e.g.,\
iterative low-pass edge-preserving filters for super-resolution of images.
Two examples of super-resolution illustrate our technology:
a no-flash RGB photo guided using a high resolution flash RGB photo, and
a~depth image guided using a high resolution RGB photo.
\end{abstract}


%
\IEEEpeerreviewmaketitle

\section{Introduction}
\label{secI}

Super-resolution (SR) refers to techniques that reconstruct high-resolution (HR) images from low-resolution (LR) images \cite{PPK03,FREM04,PETM09,M10,RIM17},
determining high-frequency components and removing degradation caused by image acquisition in LR cameras.
A~single-image edge-preserving SR by interpolation, called the zooming problem in \cite{C04,CCCNP10}, is not well-posed due to an ambiguity of determining high-frequency components, needed to preserve edges in the reconstructed high-resolution (HR) image. To remove the ambiguity and make the SR problem well-posed, one can introduce a guiding subspace, determined by frames, or via an action of an orthogonal projection on it; see, e.g., \cite{GKTM15,KGMT17}.

Frame-less guided SR using an orthogonal projector on the guiding subspace is theoretically well justified, e.g.,\ in \cite{GKTM15,KGMT17}, but may be difficult in practice, where even the guiding subspace itself may be not explicitly known. We extend the approach of \cite{GKTM15,KGMT17} to guiding operators, which increase signal components in the guiding subspace, relative to those in a~complementary subspace. As noticed in \cite{KGMT17}, the traditional Tikhonov's regularization may be substituted with pre- and post-smoothing of sample-consistent reconstruction in case of noisy samples. We adopt a similar approach and provide alternative to \cite{KGMT17} mathematical justification covering general guiding operators, not necessarily orthogonal projectors.

We illustrate our sample-consistent operator guided signal reconstruction with pre- and post-smoothing for imaging.
An HR image is reconstructed from a noisy LR image using an HR image of the same scene but in different modality as a guidance in setting up an edge-preserving denoising filter.
Several authors study a corresponding application, where the LR image is a depth image and
the HR guidance image is an RGB image of the same scene; e.g.,\ see \cite{6618873,FRRRB13,WOTV14,KB17}.

\section{Notation and prior work}
\label{secII}

A sampled (degraded) signal vector $y$ in signal processing is often represented by the linear model
\begin{equation}\label{eq1}
y = Ax+n,
\end{equation}
where the vector $x$ is the latent signal (image), $n$ is a noise vector, e.g.,\ consisting of independent and
identically distributed zero mean noise. The matrix $A$ is typically a product of
down-sampling (e.g.,\ decimation) and blurring operators.

The most frequently used restoration method is given by Tikhonov's regularization model, see, e.g.,\ \cite{M10,CCCNP10,SKM11},
\begin{equation}\label{eq2}
\min_{x}\|Ax-y\|^2_\omega+\rho R(x),
\end{equation}
where the functional $R(x)$ is called a regularization term, and $\rho>0$ is a regularization parameter.
The regularization term $R(x)$ aims at bounding and smoothing the solution $x$ of \eqref{eq2}.
The weighted norm $\|\cdot\|_{\omega}$ is defined by $\|x\|_{\omega}^2=x^T\omega x$, where
$\omega$ is a positive definite matrix. If $A=I$ is the identity matrix, then (\ref{eq2}) describes denoising.

Signal (image) processing often uses total variation due to its remarkable ability to preserve
contours/edges of signals.
The total variation term $R(x)$ approximates $\|\nabla x\|_1$; see, e.g., \cite{CCCNP10,FRRRB13,SKM11,LCBM12}.
Total variation filtering analogs can be set up in a framework of graph-based signal processing,
e.g., by setting $R(x) = x^T L(g)x$, where $L(g)$ is a graph Laplacian matrix guided by a signal $g$, e.g.,\ \cite{KM16}, that we define next.


A signal (image) is interpreted as an intensity function on $N$ vertices $V$ of a weighted graph
$G=(V,E,W)$ consisting of a finite set $V$ of vertices (e.g.,\ representing image pixels) and a finite set
$E\subset V\times V$ of edges $(i,j)$ with typically nonnegative (cf., \cite{K17,knyazev2015edge}) weights $W(i,j)$, which measure similarity between vertices $i$ and $j$ in the graph. The intensity
values of the signal form the vector $g=[g_1,\ldots,g_N]^T$, where the vertices $V$ are arbitrarily numbered.
The similarity weights form a symmetric $N\times N$ graph adjacency matrix $W=W(g)$.

Row-sums $d=W1_N$, where $1_N$ is the vector of ones, of $W$ define the diagonal $N\times N$ degree matrix $D={\rm diag}(d)$ of the graph. 
Desired for smoothing filters property $d=1_N$, i.e. $D=I$, of the graph adjacency matrix $W=W(g)$ can be ensured by scaling $W$ with positive diagonal matrix multipliers via Sinkhorn's algorithm; e.g.,\ \cite{KM14}. Alternatively, $W$ can be
substituted by the normalized graph adjacency matrix $D^{-1}W$, although the latter is technically non-symmetric.

The symmetric and positive semidefinite graph Laplacian matrix $L=L(g)$ is then defined as $L=D-W$. In graph-based signal processing, eigenvectors of $L$ serve as generalizations of basis functions of the Discrete Cosine Transform. If the guidance image $g$, used to determine $L(g)$, is aligned with an noisy image $x$ and shares the same edges, edge-preserving denoising of $x$ can be performed using filters based on $L(g)$. For example, the multiplication $W(g)x$ on $x$ amplifies spectral components in $x$ corresponding to large eigenvalues of $L$ and thus can be viewed as an approximate high-pass filter; cf. \cite{HST13}.

With $R(x) = x^T L(g)x$, the normal equations for the optimization problem \eqref{eq2} become
\begin{equation}\label{eq4}
A^T\omega(Ax-y)+\rho L(g)x=0,
\end{equation}
where $L(g)$ is the graph Laplacian operator with a guidance signal $g$.
A self-guided choice $g=x$ is nonlinear; e.g.,\ \cite{KM16}.

Graph-based interpretations of denoising filters are common.
Classical \emph{bilateral} and \emph{total variation} filters for image denoising are usually constructed via explicit formulas for the weights $W(i,j)$ on a priori determined graph edges $E$; e.g., \cite{WOTV14,KM16}. More recent \emph{guided filter} \cite{HST13} is defined directly via explicit description of its action $W(g)x$ on a given image~$x$, while its edges $E$ and formulas for the weights $W(i,j)$ are then derived for theoretical purposes only. Sophisticated filters, e.g., BM3D \cite{DFKE10}, are determined exclusively by functions that implement their action, so it may be difficult to obtain explicit formulas for their weights $W(i,j)$. Exact low-pass filters, where orthogonal projectors represent $W$ and $W1_N=1_N$, also fit the graph-based framework, although some weights $W(i,j)$ are negative.

System \eqref{eq4} with the weight matrix
$\omega = I+\beta L(g)$, where $\beta\ge-1$ is a parameter, is used in \cite{KM14}. The resulting system
\[
(A^T[I+\beta L]A+\rho L)x=A^T[I+\beta L]y
\]
is solved by the conjugate gradient (CG) method; see \cite{KM14}.

Connecting Tikhonov's regularization equation \eqref{eq2} to frame-based reconstruction,
one can set $\omega = \left(A A^T\right)^+$,  where the operation $^+$ denotes the Moore-Penrose pseudoinverse,
which gives $A^+=A^T\left(A A^T\right)^+$ and turns \eqref{eq4} into
\begin{equation}\label{eq4p}
A^+(Ax-y)+\rho L(g)x=0.
\end{equation}
Denoting $s=A^+y$ and introducing the orthogonal projector $S=A^+A$,
we equivalently rewrite equation \eqref{eq4p} as
\begin{equation}\label{eq4s}
Sx-s+\rho L(g)x=0,
\end{equation}
which is the normal equation for Tikhonov's regularization
\begin{equation}\label{eq9s}
\min_{x\in\mathbb{R}^n} \|Sx-s\|_2^2 + \rho x^TLx,
\end{equation}
---a particular case of optimization problem \eqref{eq2}.

The authors of \cite{GKTM15,KGMT17} investigate the case of problem~\eqref{eq9s},
where $L=I-W$ is an orthogonal projector. In~particular, they prove that
when the regularization parameter $\rho$ vanishes unconstrained minimization \eqref{eq9s} reduces to
\begin{equation}\label{eq5}
\min_x x^TLx\mbox{, subject to }Sx-s=0,
\end{equation}
which can be interpreted as graph-harmonic sample-consistent signal extension, since the quadratic form $ x^TLx$ can be viewed as energy of the signal $x$, defined by the graph Laplacian $L$.

Since $S$ is an orthogonal projector, we have $Sx=s=Ss$ and the orthogonal decomposition $x=s+(I-S)x$ helps to show that
the minimizer $x$ in \eqref{eq5} solves the system
\begin{equation}\label{eq6}
(I-S)L(I-S)x=-(I-S)Ls,
\end{equation}
where the matrix $(I-S)L(I-S)$ is symmetric positive semidefinite.
The special structure of equation \eqref{eq6} allows applying CG method
to the linear system
\begin{equation}\label{eq7}
(I-S)Lu=-(I-S)Ls,
\end{equation}
with an initial approximation from the null-space $\mbox{Null}(S)$ of $S$ to iteratively approximate the solution $u$ within the subspace $\mbox{Null}(S)$. The~sample-consistent reconstruction $x$ in \eqref{eq5} is then
given by the orthogonal sum $x=s+u$; see \cite{GKTM15,KGMT17} for details.

The authors of \cite{DFKE10} propose solving a self-guided nonlinear version of \eqref{eq5}
via simple iteration
\begin{equation}\label{eq8}
x_0=s,\quad x_{i+1}=s+(I-S)\mbox{BM3D}(x_i),
\end{equation}
where $\mbox{BM3D}(x_i)$ is an application of the BM3D filter to $x_i$.

Post-processing by $\alpha x + (1-\alpha) Wx$ with $\alpha = 1/(\rho+1)$ is proved in \cite{KGMT17} to solve Tikhonov's regularization problem~\eqref{eq9s}. Since $\alpha \approx 1-\rho$ for small $\rho$,
the post-processing is approximated by $x-\rho Lx$ when $\rho\to0$.
The corresponding arguments in \cite{KGMT17} rely on the assumptions that $L=I-W$ and that $W$ is an orthogonal projector.
We~provide an alternative analysis, dropping these assumptions, in the next section.

\section{Tikhonov's regularization demystified}
\label{secIII}
Equation \eqref{eq4s} implies $(I-S)Lx=0$, i.e.
\begin{equation}\label{eq6r}
(I-S)L(I-S)x=-(I-S)LSx,
\end{equation}
which differs from \eqref{eq6} only in the right-hand side,
since the sample consistency $Sx=s$ is not enforced in \eqref{eq4s}.

In an orthonormal basis of $\mathbb{R}^N$, where
\begin{equation*}
S=\begin{bmatrix}0\\&I\end{bmatrix},~
L=\begin{bmatrix}L_{11}&L_{12}\\L_{21}&L_{22}\end{bmatrix},~
x=\begin{bmatrix}x_1\\x_2\end{bmatrix},~s=\begin{bmatrix}0\\s_2\end{bmatrix},
\end{equation*}
equation \eqref{eq4s} takes the following block form,
\begin{equation}\label{eq12}
\begin{bmatrix}\rho L_{11}&\rho L_{12}\\\rho L_{21}&I+\rho L_{22}\end{bmatrix}
\begin{bmatrix}x_1\\x_2\end{bmatrix}=\begin{bmatrix}0\\s_2\end{bmatrix}.
\end{equation}
Assuming that the block $L_{11}$ is invertible, we solve the equivalent to \eqref{eq6r} equation
$L_{11}\,x_1 = - L_{12}\,x_2$ for $x_1=-L_{11}^{-1}L_{12}\,x_2$
and then for $x_2$ we obtain the following equation
\begin{equation*}
(I+\rho L/L_{11})x_2=s_2,\mbox{ where } L/L_{11}=L_{22}-L_{21}L_{11}^{-1}L_{12},
\end{equation*}
is the Schur complement of the block $L_{11}$ in the matrix $L$.

If $\rho\to0$, we conclude that $x_2\to s_2$ and $x_1\to-L_{11}^{-1}L_{12}s_2$,
so that the limit vector $x$ solves \eqref{eq5} and \eqref{eq6}, where $x_2=s_2$, i.e. $Sx=s$.
For small $\rho$, we have $x_2\approx s_2-\rho L/L_{11}\,s_2$, i.e. the solution $x$ of Tikhonov's regularization equation \eqref{eq4s} 
depends linearly on $\rho$. The minuend $L_{22}\,s_2$ in $L/L_{11}\,s_2$ corresponds to the vector $SLs$. Thus, rather than solving \eqref{eq4s} directly, for small $\rho$ one can try sample-consistent reconstruction via solving \eqref{eq5} or \eqref{eq6}, but with an a priori denoised sampled signal $s-\rho SLs$.


\section{The problem and proposed algorithms}
\label{secIV}

We are given two images of the same scene, but in different modalities.
The first image $y$ has low resolution and may be noisy.
The second image $g$ is of high resolution and is noise-free.
The images are registered (or aligned) by means of a linear downsampling transform,
that is, there is a well-conditioned matrix $A$ such that the images $Ag$ and $y$
are aligned. We want to reconstruct an image $x$ of the same resolution as $g$
from the image $y$ so that the image $g$ serves as a~guidance image in the regularization term
of the Tikhonov's regularization reconstruction model \eqref{eq4p}.

In contrast to the traditional approach directly solving \eqref{eq4p}, we use our arguments above
to take the sample-consistent reconstruction approach \eqref{eq5} from \cite{GKTM15} as a starting point,
but for a more general case where the Laplacian $L$ is an arbitrary positive semi-definite operator, not necessarily an orthogonal projector. We notice that the proposed in \cite{GKTM15,KGMT17} reduction of \eqref{eq5} to \eqref{eq7} also works for this general case, so we solve \eqref{eq7}.

Noisy LR samples $y$ should evidently be smoothed prior to computing the sample-consistent HR reconstruction $x$ by \eqref{eq7}.
Dealing with noise, we always pre-process $y$ via denoising, by analogy with $y-\rho ALA^+y$ proposed in Section~\ref{secIII}. We~also find it helpful in some tests to post-process our sample-consistent reconstruction $x$,
similar to subtracting $x-\rho Lx$ highly-oscillatory contributions described in Section~\ref{secII}.

%

Let $CG_m(\mathcal{A},v)$ denote a function, which implements $m$ iterations of CG to 
solve the equation $\mathcal{A}(u)=v$ with a given linear operator $\mathcal{A}(u)$. 
The main cost per iteration is the cost of evaluation of $\mathcal{A}(u)$. 
Our reconstruction method is as follows.

\vspace{1ex}
\begin{tabular}{l}
\hline\\[-2ex]
\textbf{Algorithm 1}{ Operator guided super-resolution} by CG\\\hline\\[-2ex]
\textbf{Input:} sample $y$, downsampling operator $A$, \\guiding operator $L$, number of iterations $m$.\\
\textbf{Output:} super-resolution reconstruction $x$ of $y$:\\
Define operator $\mathcal{A}(u)=(I-A^+A)Lu$.\\
Choose an initial approximation $x^0$ satisfying $Ax^0=y$.\\
Compute $x = x^0 - CG_m(\mathcal{A},\mathcal{A}(x^0))$.\\
\hline\\[-2ex]
\end{tabular}
\vspace{1ex}
 
The level  $x^T1_N$ of the DC-component $1_N$ in the reconstructed HR signal $x$ can be adjusted during post-processing to match that in the LR sample $y$, if necessary.


%

Compared to \cite{GKTM15}, where $W$ and $L=I-W$ are assumed to be orthogonal projectors, our approach allows choosing more general smoothing filters $W$, but there are still limitations:
\begin{itemize}
\item Negative entries in the filter matrix $W$ are allowed, but the resulting Laplacian $L=D-W$
    needs to be symmetric positive semidefinite, in order for the minimization of its quadratic form $x^TLx$ in \eqref{eq5} to make sense, and to satisfy assumptions for CG convergence in Algorithm 1.
\item DC-invariant filters, i.e. satisfying $W1_N=1_N$, lead to $D=I$ and thus to the trivial construction of $L=I-W$, 
such as, e.g.,\ the image guided filter of \cite{HST13}, implemented in the MATLAB Image Processing Toolbox, and the total variation filter presented in \cite{KM16}. Otherwise, the filter DC-component evaluation $d=W1_N$ is needed to determine $D=\mbox{diag}(d)$ in $L=D-W$.
\item Iterative filters are allowed, e.g., a DC-invariant smoothing filter $W$, satisfying $W1_N=1_N$, generates the DC-invariant polynomial filter $P_n(I-W)$ where $P_n(\cdot)$ is a polynomial of degree $n$, satisfying $P_n(1)=1$, for example, $P_2(W)=W^2$. However, the polynomial $P_n(\cdot)$ needs to remain fixed during the CG iterations.
\item Self-guided filters lead to nonlinear operators $\mathcal{A}(\cdot)$, requiring special care, such as proposed in \cite{KM16}.
\end{itemize}

\section{Numerical experiments}
\label{secV}

\paragraph{Super-resolution for flash-no~flash images}

\begin{figure}[htb]
\begin{minipage}[b]{1.0\linewidth}
\centerline{\includegraphics[width=.98\linewidth]{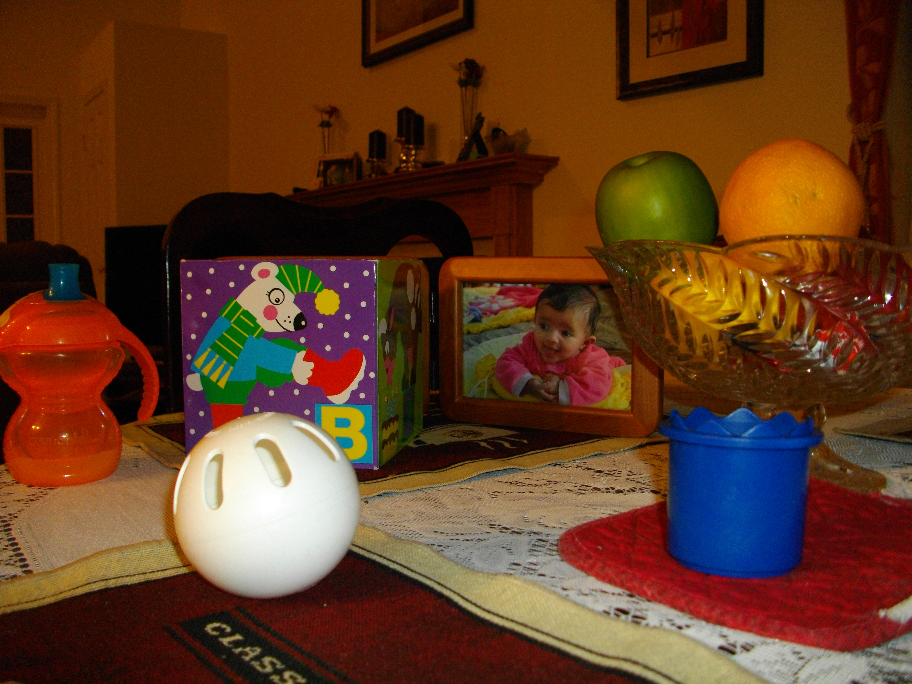}}
\end{minipage}
\caption{HR guidance image {\em toysflash}.}
\label{fig1}
\end{figure}

\begin{figure}[htb]
\begin{minipage}[b]{1.0\linewidth}
\centerline{\includegraphics[width=.98\linewidth]{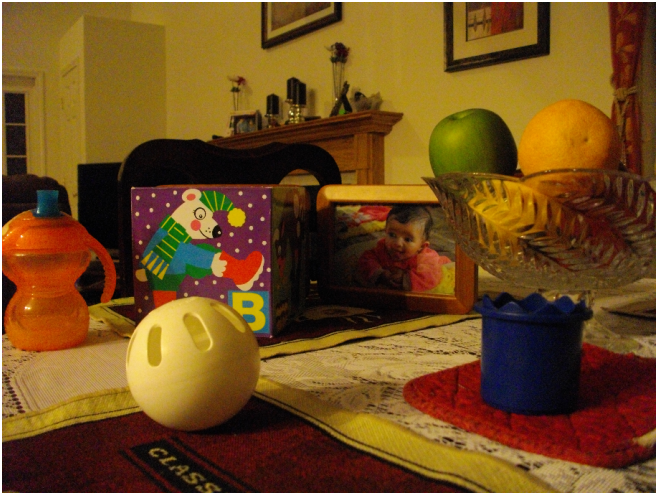}}
\end{minipage}
\caption{HR ground truth image {\em toysnoflash}.}
\label{fig2}
\end{figure}

We have carried out numerical tests in MATLAB with two registered color images from the MATLAB own directory, \texttt{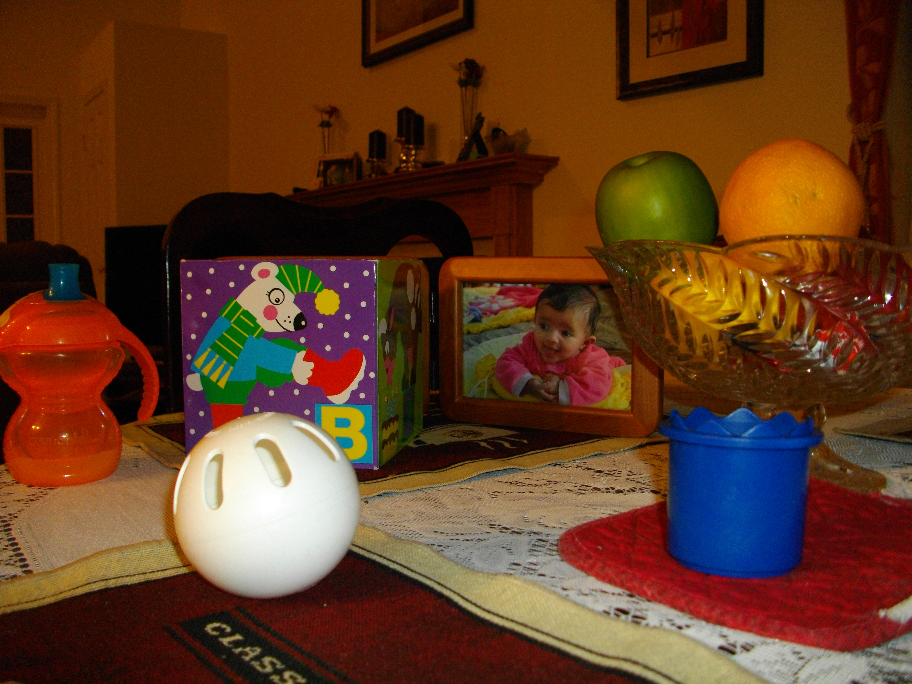} and \texttt{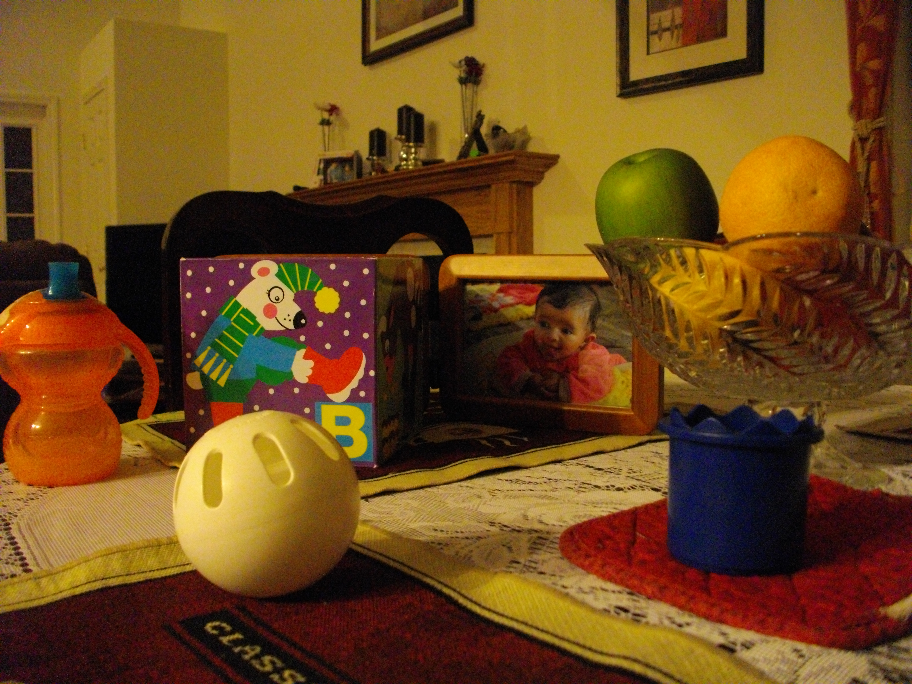} of the size $684\times912$, taken respectively
with and without flash. The first image is our HR guidance image.
We downsample the second image by factor $4$ in both dimensions choosing every $4$-th pixel with the command \texttt{image(1:4:end,1:4:end)}.
The result of downsampling is our LR image. Both images are scaled so that their intensities
lie in the range $[0,1]$. Figures~\ref{fig1} and \ref{fig2} display the guidance and the ground truth images.
The two images of the same scene have different modalities owing to big differences in color representations
of the objects. For~example, the yellow ball looks white in the flash image.

\begin{figure}[htb]
\begin{minipage}[b]{1.0\linewidth}
 \centerline{\includegraphics[width=.98\linewidth]{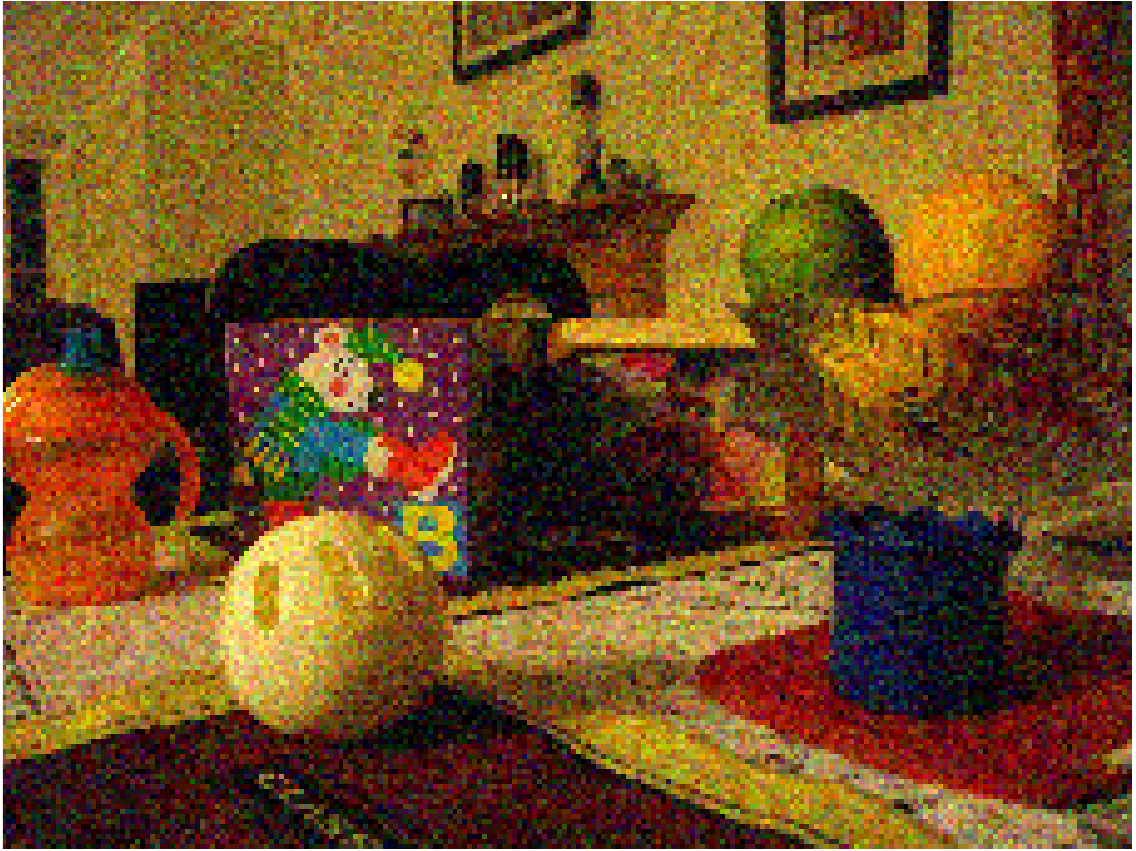}}
\end{minipage}
\caption{Noisy LR image.}
\label{fig3}
\end{figure}

To produce a noisy downsampled LR image, a Gaussian noise with the default parameters,
zero mean and variance $0.01$, has been added to the LR image. The degraded image
is displayed in Figure~\ref{fig3}. We pre-smooth the noisy LR image by the image guided
filter with the default parameters and then apply Algorithm 1 to the pre-smoothed image.

We use a single application of the guided filter function \texttt{imguidedfilter} from the MATLAB Image Processing Toolbox as the smoothing filter $W(g)$ for color images.
The~function parameters are as follows: width of neighborhoods is $7$, the smoothing value is $10^{-6}$.

\begin{figure}[htb]
\begin{minipage}[b]{1.0\linewidth}
 \centerline{\includegraphics[width=.98\linewidth]{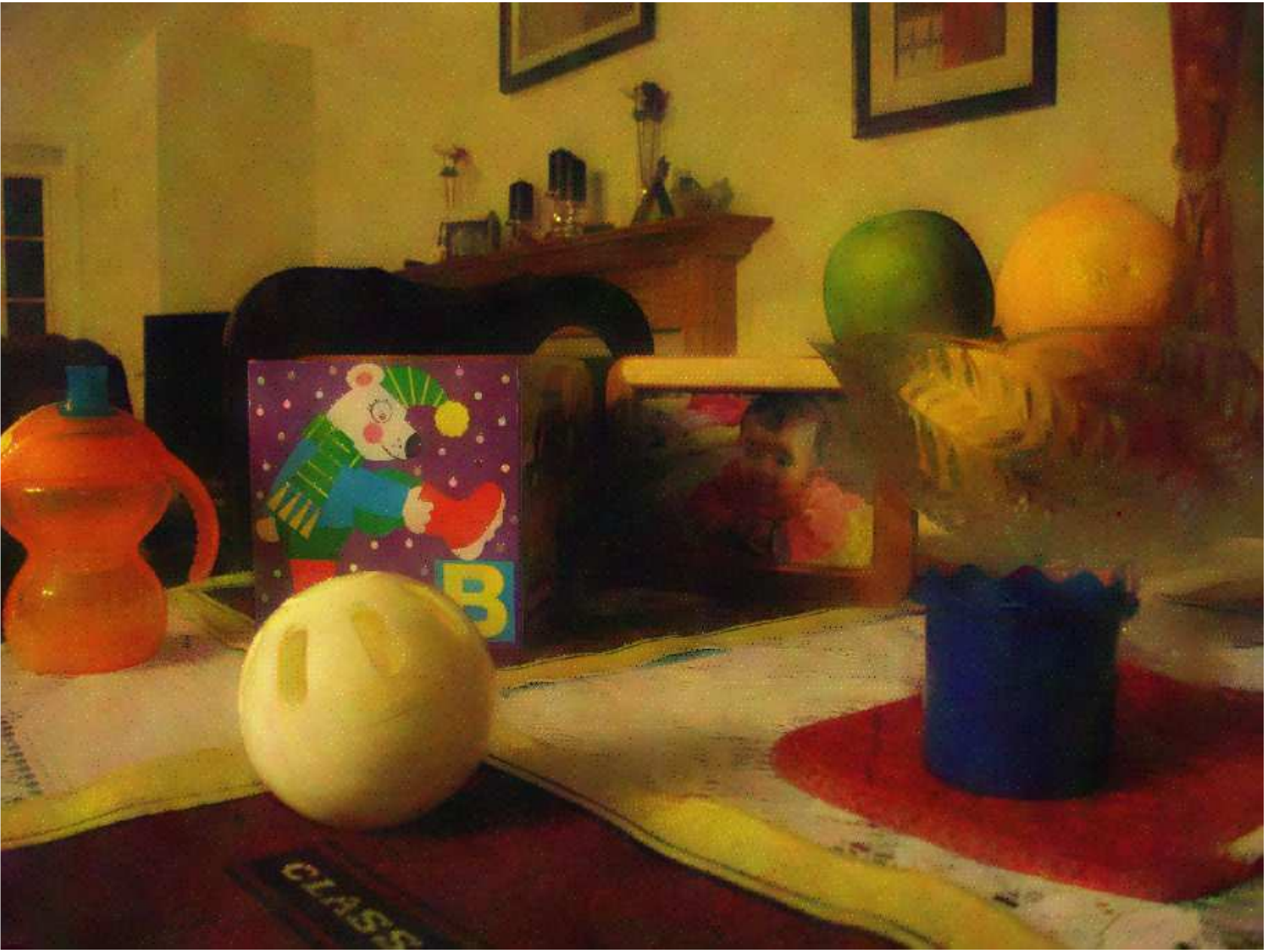}}
\end{minipage}
\caption{Our reconstructed image, PSNR = 24.32.}
\label{fig4}
\end{figure}

The relative residual is $10^{-3}$ after $m=20$ CG iterations, 
which translates into $20$ applications of \texttt{imguidedfilter}.
Figure~\ref{fig4} shows the reconstructed image, which has the same desired colors as
the ground truth image. The PSNR referred to the ground truth HR image equals $24.32$.

\paragraph{Super-resolution of a depth image}

\begin{figure}[htb]
\begin{minipage}[b]{1.0\linewidth}
 \centerline{\includegraphics[width=.98\linewidth]{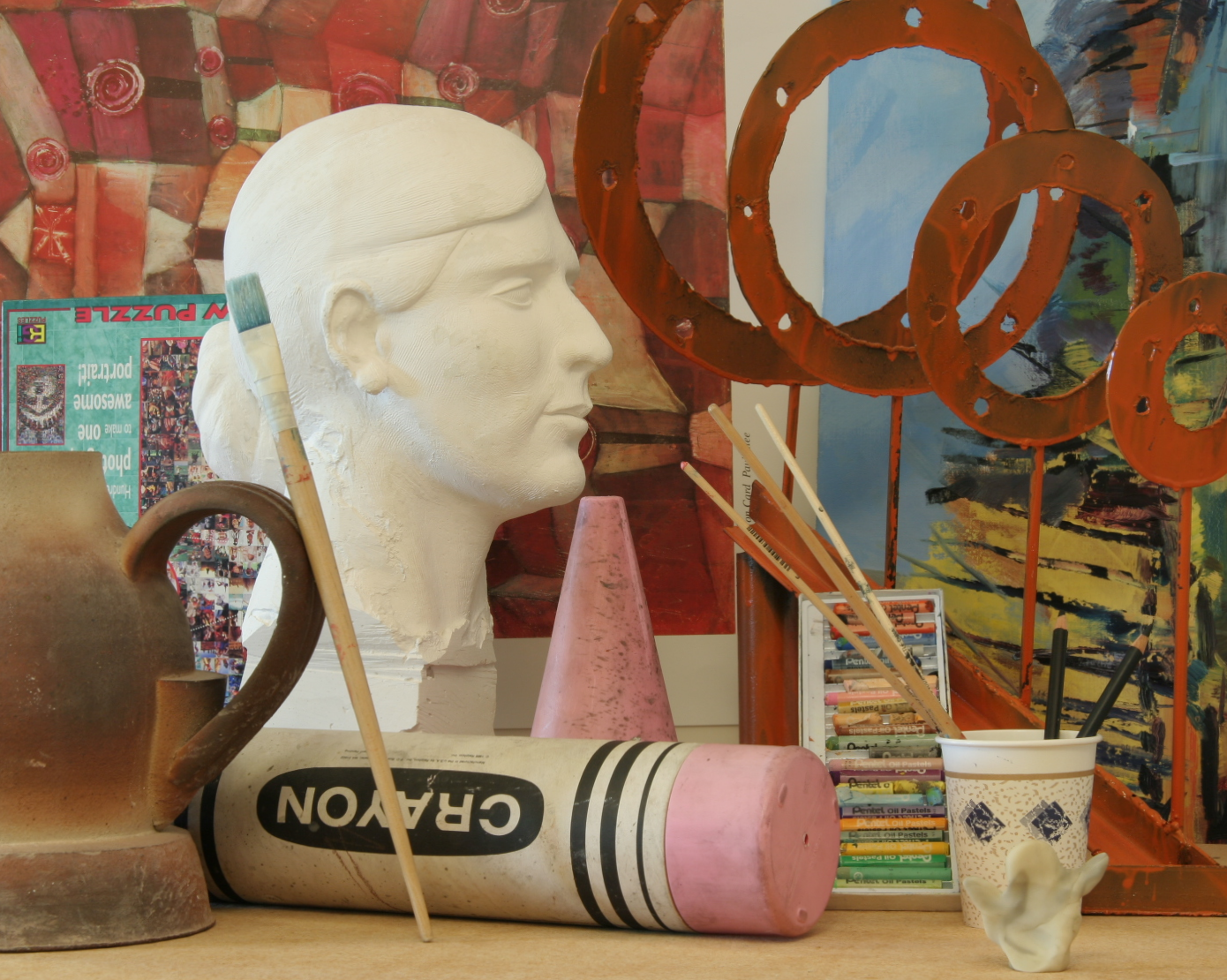}}
\end{minipage}
\caption{High resolution RGB $1390\times1110$ image \emph{Art}.}
\label{fig:art_rgb}
\end{figure}

\begin{figure}[htb]
\begin{minipage}[b]{1.0\linewidth}
 \centerline{\includegraphics[width=.98\linewidth]{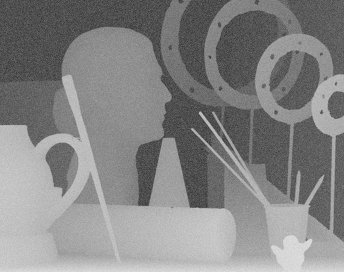}}
\end{minipage}
\caption{Low resolution noisy depth $344\times272$ image \emph{Art}.}
\label{fig:art_depth_2_n}
\end{figure}

\begin{figure}[htb]
\begin{minipage}[b]{1.0\linewidth}
 \centerline{\includegraphics[width=.98\linewidth]{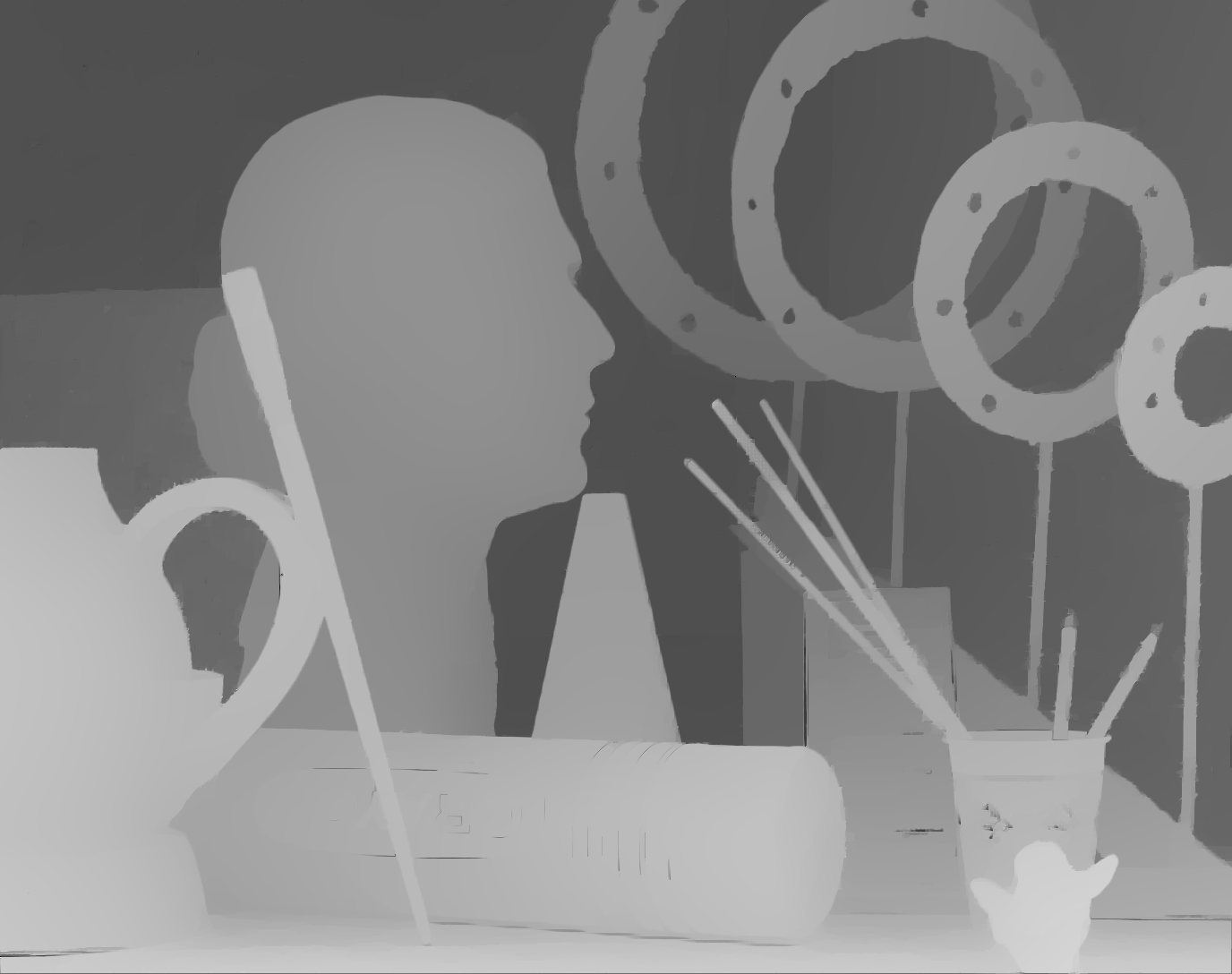}}
\end{minipage}
\caption{Upsampled depth $1376\times1088$ image \emph{Art}, using \cite{FRRRB13}. PSNR = $32.8$.}
\label{fig:art_FRRRB}
\end{figure}

Algorithm 1 has been also evaluated on the \emph{Art} image set from the simulated Middlebury 2007 data sets extensively used in \cite{FRRRB13}. The~guidance image is the gray component of the HR RGB image \emph{Art} in Figure~\ref{fig:art_rgb}. The LR noisy depth image is shown in Figure~\ref{fig:art_depth_2_n}. The upsampling shown in Figure~\ref{fig:art_FRRRB} has been computed by the code used in \cite{FRRRB13}, which implements
Tikhonov's regularization by the generalized total variation.

\begin{figure}[htb]
\begin{minipage}[b]{1.0\linewidth}
\vspace*{0.6ex}
 \centerline{\includegraphics[width=.98\linewidth]{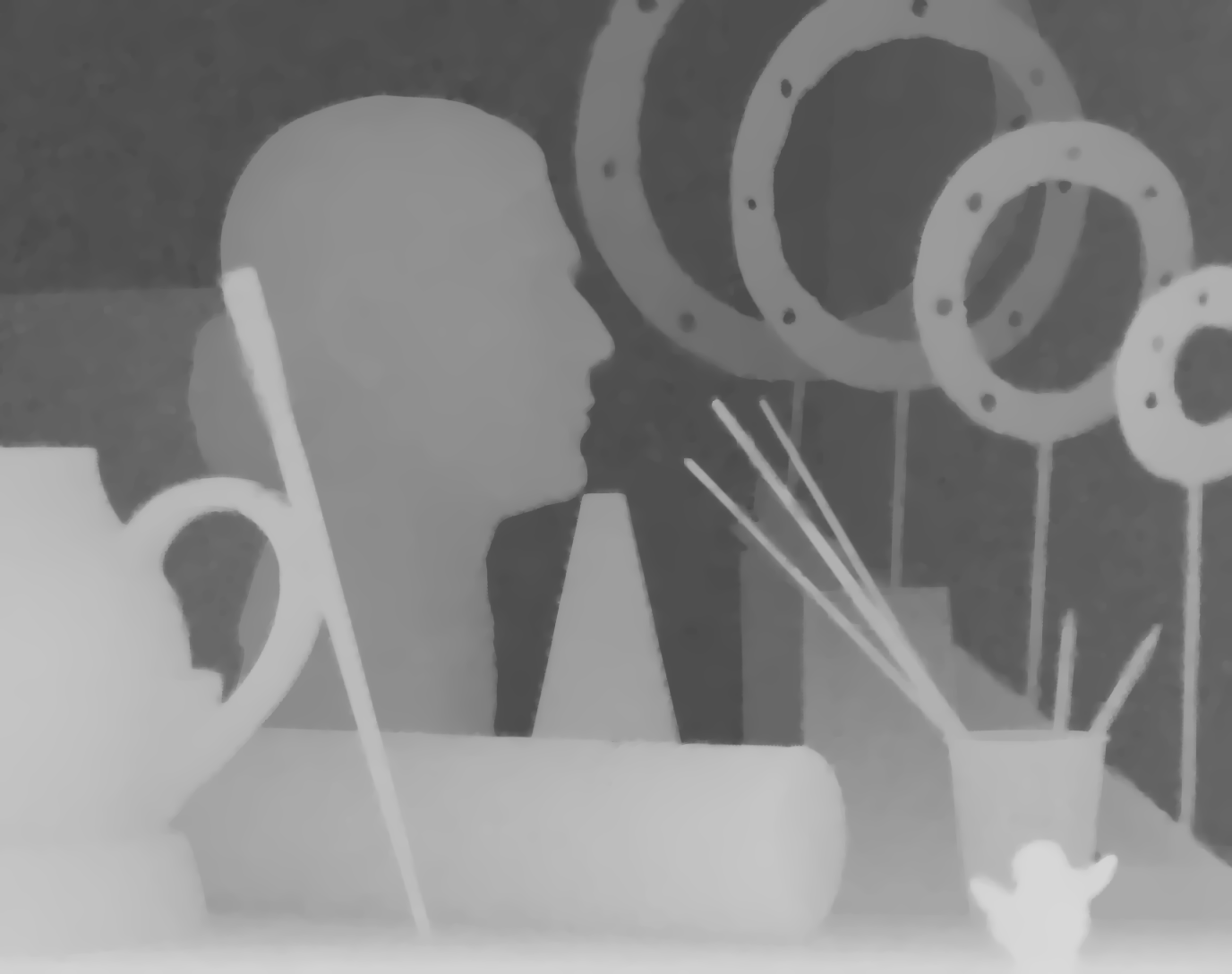}}
\end{minipage}
\caption{Upsampled depth $1376\times1088$ image \emph{Art}. Our result. PSNR = $33.6$.}
\label{fig:art_interp}
\end{figure}

To produce Figure~\ref{fig:art_interp} we use the guided Total Variation (TV) filter described in \cite{KM16}
as the guided smoothing filter in Algorithm~1 as well as during LR pre-smoothing and HR post-smoothing. 
The smoothing parameter used is $10^{-4}$ at all three stages, i.e. inside Algorithm~1, as well as for 
the pre- (and post-) smoothing, with 1/480/850 function evaluations, correspondingly.
The number $m$ of CG iterations is $2900$ to achieve the relative residual $2\cdot10^{-14}$.

Comparison Figure \ref{fig:art_FRRRB} has a smaller PSNR, but a bit less noisy and has sharper edges, than our Figure \ref{fig:art_interp}, 
as the generalized TV filter used in \cite{FRRRB13} is more powerful smoother compared to the TV filter from \cite{KM16} 
 that we use in these tests. However, Figure~\ref{fig:art_FRRRB} has noticeable artifacts, clearly coming from
sharp edges in RGB Figure \ref{fig:art_rgb}, which are absent in LR Figure \ref{fig:art_depth_2_n}. In our approach,
sharp edges in the guidance image $g$ only affect the filter weights and thus cannot possibly pollute our reconstruction, as confirmed in Figure \ref{fig:art_interp}.

\section{Conclusion}

We propose algorithms for guided iterative reconstruction of signals, illustrated with reconstruction of a higher resolution image from a lower resolution noisy sample. Guiding is given by a smoothing filter and plays the role of regularization. Least squares minimization of the data term $\|Ax-y\|_2$ is substituted by the equality constraint $Ax=y$, allowing one to control sample consistency and eliminating the problem of choosing the regularization parameter in Tikhonov's regularization.

Noisy lower resolution samples are pre-smoothed. The obtained sample-consistent high resolution reconstruction can be post-smoothed, if needed. Iterations are based on conjugate gradients, giving the optimal performance.
General smoothing guidance filters allow flexibility in designing reconstructions with desirable properties.
Initial numerical experiments demonstrate feasibility of the proposed technology
for super-resolution for flash-no~flash images and for depth reconstruction guided by RGB images.

\bibliographystyle{IEEEbib}
\bibliography{refs}

\end{document}